\documentclass[sigconf]{acmart}
\AtBeginDocument{%
  }

\setcopyright{acmcopyright}
\copyrightyear{2018}
\acmYear{2018}
\acmDOI{XXXXXXX.XXXXXXX}

\acmPrice{15.00}
\acmISBN{978-1-4503-XXXX-X/18/06}

\usepackage{bm}
\usepackage{multirow}
\usepackage{array}
\usepackage{arydshln} % for cdashline
\newcommand{\tabincell}[2]{\begin{tabular}{@{}#1@{}}#2\end{tabular}}
\usepackage{pifont}
\usepackage{subfigure}  % sub figure in the same line

\usepackage{color}
\newcommand{\sst}[1]{{\color{black}#1}}
\newcommand{\sgg}[1]{{\color{black}#1}}
\usepackage{caption}

\acmSubmissionID{2492}

\renewcommand\footnotetextcopyrightpermission[1]{}
\begin{document}

%\title{Exploiting Mid-Level Semantic Knowledge for Federated
%Learning}
% \title{\sgg{Exploring Semantic Attributes from A Language Foundation Model
%   for Federated Zero-Shot Learning of Disjoint Label Spaces}}
\title{\sgg{Exploring Semantic Attributes from A Foundation Model
for Federated Learning of Disjoint Label Spaces}}
% \author{Shitong Sun}
% % \authornote{Both authors contributed equally to this research.}
% \email{shitong.sun@qmul.ac.uk}
% % \orcid{1234-5678-9012}
% \affiliation{%

\author{Shitong Sun}
% \affiliation{%
%   \institution{Queen Mary University of London}
% %   % % \streetaddress{1 Th{\o}rv{\"a}ld Circle}
% %   % % \city{London}
%   \country{United Kindom}
%   }
\email{shitong.sun@qmul.ac.uk}

\author{Chenyang Si}
% \affiliation{%
%   \institution{Institute of Automation, Chinese Academy of Sciences}
% %   \city{Rocquencourt}
%   \country{China}
% }
\email{chenyang.si.mail@gmail.com}

\author{Guile Wu}
% \affiliation{%
\email{guile.wu@outlook.com}

\author{Shaogang Gong}
% \affiliation{%
%   \institution{Queen Mary University of London}
% %   % % \streetaddress{1 Th{\o}rv{\"a}ld Circle}
% %   % % \city{London}
%   \country{United Kindom}
%   }
\email{s.gong@qmul.ac.uk}

\renewcommand{\shortauthors}{Trovato et al.}

\begin{abstract}
Conventional centralised deep learning paradigms are not feasible when data from different sources cannot be shared due to data privacy or transmission limitation.
To resolve this problem, federated learning has been introduced to transfer knowledge across multiple sources (clients) with non-shared data while optimising a globally generalised central model (server).
Existing federated learning paradigms mostly focus on transferring holistic high-level knowledge (such as class) across models, which 
% can result to weight divergence without independent and identically discributed (IID) samples across local clients. 
is closely related to specific objects of interest so may suffer from inverse attack.
In contrast, in this work, we consider transferring mid-level semantic knowledge (such as attribute) which is not sensitive to specific objects of interest and therefore is more privacy-preserving and general.
% generic containing semantically meaningful properties among clients even in non-IID setting.
%
To this end, we formulate a new Federated Zero-Shot Learning (FZSL) paradigm to learn mid-level semantic knowledge at multiple local clients with non-shared local data and cumulatively aggregate a globally generalised central model for deployment.
To improve model discriminative ability, we \sgg{explore semantic
  knowledge} \sst{\sgg{available from either a language or a vision-language foundation
    model in order} to enrich} the mid-level semantic space in FZSL.
Extensive experiments on five zero-shot learning benchmark datasets validate the effectiveness of our approach for optimising a generalisable federated learning model with mid-level semantic knowledge transfer.
\end{abstract}


% \ccsdesc[500]{Computing methodologies~Transfer learning}




\maketitle

\begin{figure}[t]
  \centering
  \includegraphics[width=0.5\textwidth]{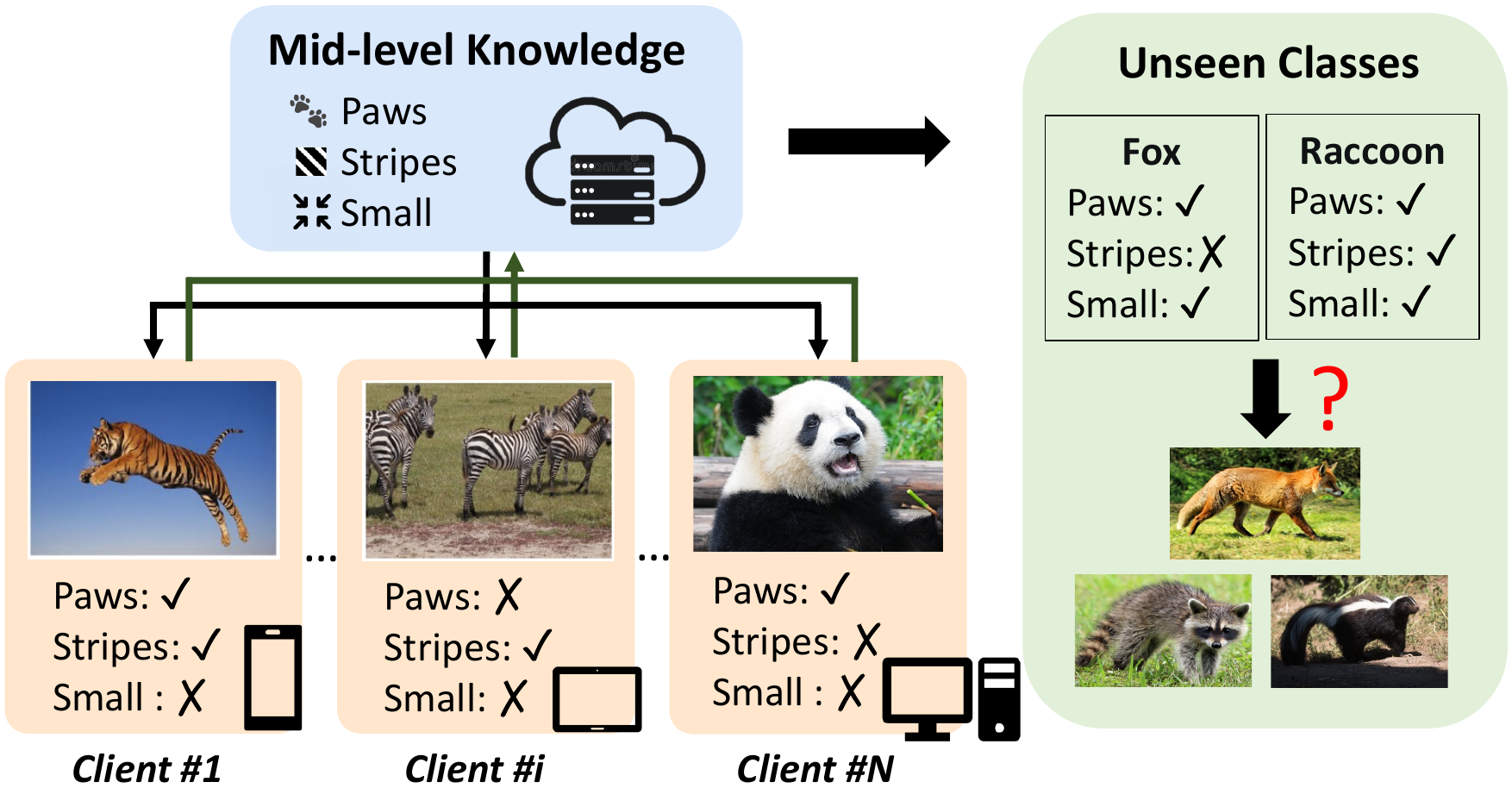}
  \caption{An overview of federated zero-shot learning with mid-level semantic knowledge transfer.
  Each local client optimises a local model with non-shared local data whilst a central server aggregates a global model by aggregating local model parameters.
  The server model will further be tested on unseen novel classes.}
  \label{fig:overview}
\end{figure}

\section{Introduction}
\label{sec:intro}
Deep learning has gained great success in computer vision and natural language processing, but conventional deep learning paradigms mostly follow a centralised learning manner where data from different sources are collected to create a central database for model learning.
With an increasing awareness of data privacy, decentralised deep learning~\cite{mcmahan2017communication,wu2021decentralised} is more desirable.
To this end, federated learning~\cite{mcmahan2017communication,li2020fedbn} has been recently introduced to optimise local models (clients) with non-shared local data while learning a global generalised central model (server) by transferring knowledge across the clients and the server.
This enables to protect data privacy and reduce transmission cost as local data are only used for training local models and only model parameters are transmitted across the clients and server.
There have been a variety of federated learning methods for computer vision applications, such as image classification~\cite{chen2021bridging}, person reidentification~\cite{sun2021decentralised} and object detection~\cite{liu2020fedvision}.

However, existing federated learning methods~\cite{mcmahan2017communication,li2020fedbn,wu2021decentralised,chen2021bridging} mostly focus on learning holistic high-level knowledge transfer across the clients and the server.
%
% In case of non-IID federated learning, the training process may vary significantly according to data distribution~\cite{li2020fedbn} and even lead to weight divergence~\cite{zhao2018federated} when aiming to achieve a global server model by only holistic high-level knowledge transfer, especially when there is even non-overlapping class space among clients. 
%
%
Since high-level knowledge is closely related to objects of interest, this may be threatened by privacy inference attacks aiming to reconstruct the raw data ~\cite{liu2022threats}.
% this may pose a threat to data privacy.
%
In contrast, mid-level semantic knowledge (such as attribute) is generic containing semantically meaningful properties for visual recognition~\cite{lampert2013attribute} and
% , which can be the general knowledge extracted from heterogenous clients.  
not sensitive to objects of interest, so the attacks to reconstruct the sample label space through gradient leakage~\cite{zhu2019deep,zhao2020idlg,geiping2020inverting} can be avoided. 
Besides, since the number of attributes is finite in compositional learning~\cite{yuille2021deep} but the number of classes can be infinite, we can represent many classes by the same dictionary of attributes~\cite{yuille2011towards}, similar to human cognitive abilities~\cite{yuille2021deep,lampert2013attribute} and generalisation requirements for the server model in federated learning.
% mid-level knowledge is also supposed to be more scalable.
%data efficient (than if we have to learn each object separately)~\cite{yuille2011towards}. 
%
Therefore, learning mid-level semantic knowledge transfer for federated learning is not only critial but also desirable for protecting privacy as well as improving model generalisation.
%heterogenous federated learning is important and is desirable for boosting model generalisation and improving data efficiency.

On the other hand, zero-shot learning (ZSL) is a well-established paradigm for learning mid-level knowledge.
It aims to learn mid-level semantic mapping between image features and text labels (typically attributes) using seen object categories and then transfer knowledge for recognising unseen object categories with the help of the composition of shared attributes between seen and unseen categories.
However, existing ZSL methods~\cite{pourpanah2020review,chen2021knowledge,chen2021free} mostly consider centralised learning scenarios which require to share training data from different label spaces to a central data collection.
Thus, learning mid-level knowledge in ZSL under a decentralised learning paradigm remains an open question.

\begin{figure*}[t]
    \centering
    \includegraphics[width=0.97\textwidth]{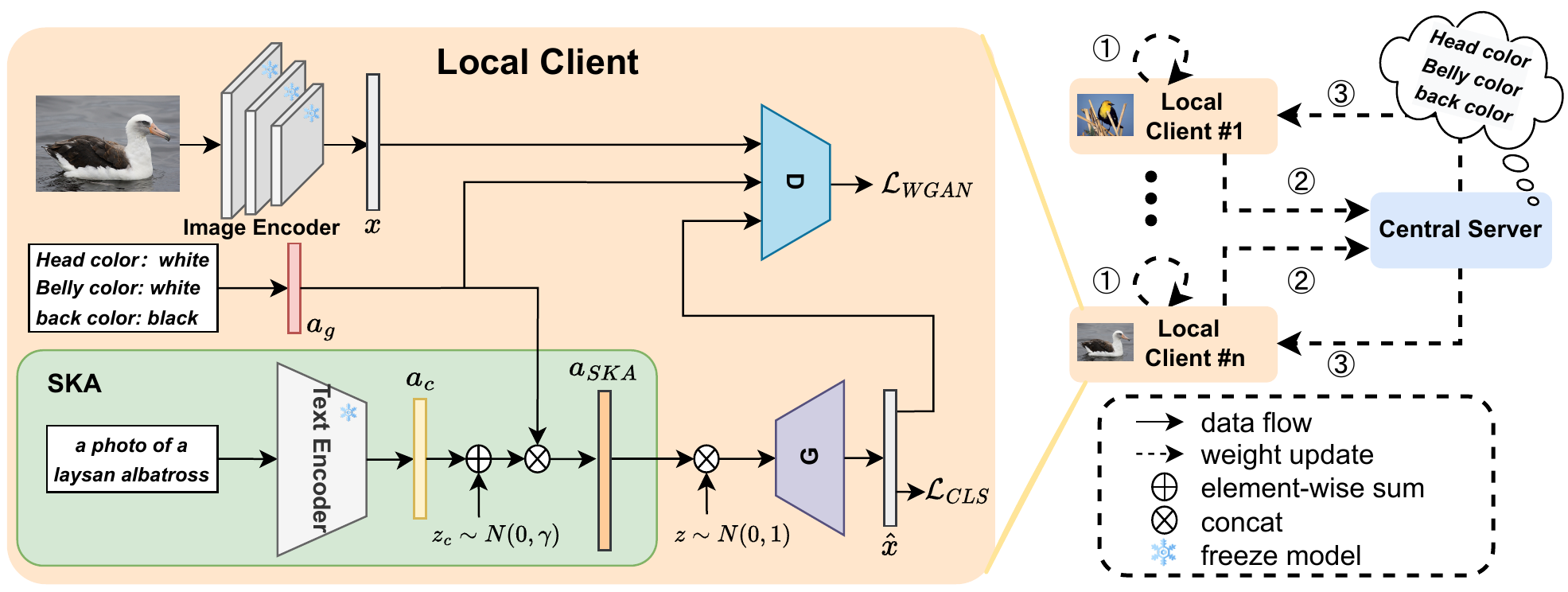}
    \caption{An overview of federated zero-shot learning with mid-level semantic knowledge transfer.
    %
  %   Each local client optimises a local model with non-shared local data whilst a central server aggregates a global model by aggregating local model parameters.
    %
    (1) Local model training process. 
    (2) Local clients upload model parameters to the server and the server constructs a global model by aggregating local model parameters.
    (3) Local models are reinitialised with the central server model. 
    The Semantic Knowledge Augmentation (SKA) employs external knowledge to further improve the model's discriminative ability.}
    \label{fig:local_client}
  \end{figure*}

  In this work, we formulate a new Federated Zero-Shot Learning (FZSL) paradigm, which aims to learn mid-level semantic knowledge in federated learning for zero-shot learning in a decentralised learning manner.
  An overview of FZSL is depicted in Fig.~\ref{fig:overview}.
  Specifically, we consider there are multiple local clients where each client has an independent non-overlapping class label space whilst all clients share a common mid-level attribute space.
  \sst{This is an extremely non-IID data partition with a high degree of label distribution skewness and mean Kolmogorov-Smirnov value of one~\cite{qu2022rethinking}.}
  Then, we optimise local models (clients) with non-shared local data and learn a central generalised model (server) by transferring knowledges (model parameters) between the clients and the server.
  Further, the server model is tested on unseen novel classes utilising the learned mid-level knowledge. 
  %
  % With this paradigm, FZSL unifies federated learning and zero-shot learning for learning mid-level semantic knowledge in a decentralised learning manner with data privacy protection.
  %
  With this paradigm, we innovatively learn mid-level semantic knowledge in a decentralised learning manner with data privacy protection by bridging federated learning and zero-shot learning.
  %non-IID data distribution.
  %non-overlapping class distribution among clients. 
  %
  It cumulatively optimises a generic mid-level attribute space from non-sharable distributed local data of different object categories.
  Instead of aggregating holistic models like traditional federated learning~\cite{mcmahan2017communication} or separating domain-specific classifiers like recent decentralized learning~\cite{wu2021collaborative,wu2021decentralised}, we only aggregate generators across the clients and the server while discriminators are retained locally.
  This facilitates to learn more generalised knowledge and reduce the number of model parameters for communicating.
  Furthermore, to improve model discriminative ability, we utilise external knowledge by employing off-the-shelf foundation model (language model (e.g., RoBERTa~\cite{liu2019roberta}) or text encoder of vision-language model (e.g., CLIP~\cite{radford2021learning}))
  to explore semantic knowledge augmentation to enrich the mid-level semantic space in FZSL.
  The text encoder is frozen in local clients and decoupled from the client-server aggregation process. 
  With the help of the pre-trained text encoder supplying richer knowledge space, this semantic knowledge augmentation allows to learn a more generic knowledge to encode sample diversity as well as model scalability.

  Our \textbf{contributions} are:
  % (1) We introduce a new Federated Zero-Shot Learning paradigm to transfer mid-level knowledge from independent non-overlapping class label spaces for federated learning;
  \sst{(1) \sgg{We} propose to exploit mid-level semantic knowledge transfer
     for federated learning from independent non-overlapping class label spaces and introduce a new Federated Zero-Shot Learning paradigm.
  % \sst{(1) We innovatively exploit mid-level semantic knowledge across non-overlapping class label spaces for federated learning in our proposed Federated Zero-Shot Learning paradigm.}
  %
  (2) \sgg{We formulate a baseline model, from which we further explore}
    semantic knowledge augmentation \sgg{by using either a language or a}
    \sst{vision-language foundation model} as external knowledge to
    learn a richer mid-level semantic space in FZSL;} 
  (3) We conduct extensive experiments on five zero-shot learning
  benchmark datasets \sgg{to} demonstrate \sgg{the capacity of our
    approach in} learning a generalised federated learning model with
  mid-level semantic knowledge transfer.
  % generalised model in non-IID federated learning with mid-level semantic knowledge transfer.

  \section{Related Work}
  \subsection{Federated Learning}
  Federated learning~\cite{mcmahan2017communication,li2020fedbn,li2020federated} is a recently introduced model learning paradigm aiming to learn a central model (server) with the collaboration of multiple local models (clients) under data privacy protection.
  It has been explored in many computer vision tasks, such as medical image segmentation~\cite{liu2021feddg}, person reidentification~\cite{wu2021decentralised}, object detection~\cite{liu2020fedvision}, etc.
  Conventional federated learning approaches, e.g., FedAvg~\cite{mcmahan2017communication}, learn a sharable central model by aggregating holistic model parameters among different local models.
  To disentangle generic and specific knowledge, recent approaches~\cite{wu2021collaborative,zhang2021fedzkt,wu2021fedcg,sun2021decentralised} propose to optimise generic feature extractors or generators by decoupling discriminators or domain-specific classifiers, but are still learning holistic class-level knowledge.
  Different from existing works, we propose to learn mid-level semantic knowledge (i.e., attributes) for federated zero-shot learning.

  \sst{Although there have been several seemingly related federated zero-shot learning studies~\cite{gudur2021zero,hao2021towards,zhang2021fedzkt,chen2022federated}, they mostly use a different definition and setting of \emph{zero-shot} and none of these methods are aimed at bridging the gap between seen and unseen classes by learning mid-level semantic knowledge.
ZSDG~\cite{hao2021towards} implements zero-shot augmentation to generate existing categories and study a sharing class space.
  FedZKT~\cite{zhang2021fedzkt} and Gudur et al.~\cite{gudur2021zero} are based on zero-shot knowledge distillation~\cite{nayak2019zero} with the purpose of transferring knowledge between clients and server without additional datasets.
  While our work is based on traditional zero shot learning task aimed to learn transferrable \emph{mid-level} semantic attributes.
  Further, our FZSL is generalisable on \emph{unseen} classes, while FedZKT and ZSDG are only tested on seen classes;  our FZSL is learning from multiple independent \emph{non-overlapping} class label spaces, while ZSDG~\cite{hao2021towards} and Gudur et al.~\cite{gudur2021zero} are studying a sharing class space.}
  Among all, FedZSL~\cite{chen2022federated} is the most similar to us, but it is based on class-level knowledge like other methods~\cite{gudur2021zero,hao2021towards,zhang2021fedzkt} 
  %
  %
  %
  % Further, our FZSL is generalisable on \emph{unseen} classes, while FedZKT and ZSDG are only tested on seen classes.
  %
  % More importantly, all of these methods are based on class-level knowledge 
  while our FZSL learns to transfer mid-level semantic knowledge.
  % Besides, we propose semantic knowledge augmentation from external knowledge to improve model discriminative ability for FZSL.

\subsection{Zero Shot Learning}
Zero shot learning (ZSL) aims to recognise unseen object categories leveraging seen categories for learning consistent semantic information to bridge seen and unseen categories. 
Current ZSL methods can broadly be divided into embedding based methods~\cite{fu2015transductive} and generative based methods~\cite{xian2018feature}.
Embedding based methods transfer from a visual space to a semantic space and classify unseen categories based on semantic similarity without any training data. 
In contrast, generative based methods learn a projection from a semantic space to a visual space, which enables to turn the zero shot learning task to a pseudo feature  supervised learning task, alleviating overfitting~\cite{xian2018feature}. 
Existing ZSL methods are following a centralised learning manner, while our work proposes a new federated zero-shot learning paradigm to transfer mid-level knowledge across different non-overlapping class label spaces with data privacy protection.

\subsection{Foundation Models}
Foundation models refer to models trained with a vast quantity of data and can be further used for various downstream tasks, such as BERT~\cite{devlin2018bert}, RoBERTa~\cite{liu2019roberta}, CLIP~\cite{radford2021learning}, etc.
These models are usually learned by self-learning using unlabelled data and can predict underlying properties such as attributes, so they are scalable and potentially more useful than models trained on a limited label space~\cite{bommasani2021opportunities}.
A concurrent work, PromptFL~\cite{guo2022promptfl}, has recently adapted a foundation model in federated learning to explore the potential representational power of the pretrained richer knowledge. 
However, it adapts the whole vision-language model (both image and text encoder) for model learning and focuses on general image classification.
\sst{\sgg{Critically}, the image encoder of the foundation models cannot be
  used \sgg{directly} in zero shot learning image
  classification~\cite{lampert2013attribute}. \sgg{This is because
    that the ZSL hypothesis of disjoint training and test
    class labels~\cite{xian2018zero,lampert2013attribute} is not guaranteed
    in vision-language foundation models}.
\sgg{To overcome this problem}, we employ \sgg{either} a language model
% (e.g., RoBERTa~\cite{liu2019roberta})
or a text encoder of a vision-language foundation model
% (e.g., CLIP~\cite{radford2021learning})
to \sgg{augment \em{only}} the mid-level semantic space in FZSL.
And we use the ImageNet as pretrain dataset and ensure non-overlapping of training and test classes by~\cite{xian2018zero}.}

\section{Methodology}
\subsection{Problem Definition}
In this work, we study Federated Zero-Shot Learning (FZSL), where each client contains an independent non-overlapping class label space with non-shared local data while a central model is aggregated for deployment.
Suppose there are $N$ local clients, where the $i$-th client contains a training set $\mathcal{S}_{i} = \left\{\bm{x}, y\right\}$, here $y \in \mathcal{Y}_{i}$ includes $N_{i}$ classes.
Since each client contains non-overlapping class space, i.e., $\{\mathcal{Y}_{i}\cap \mathcal{Y}_{j}{=}\emptyset, \forall i,j\}$, $\mathcal{Y}_{1} \cup\mathcal{Y}_{2} \cup \ldots \cup \mathcal{Y}_{N}=\mathcal{Y}_{s}$.
Meanwhile, each class can be described by an attribute vector $\bm{a}=\left\{a_1,a_2\ldots a_m\right\}$ and these $m$ attributes are consistent among classes in all clients, i.e. the mid-level attribute space is shared across clients.
The goal of federated zero shot learning task is to construct a classifier $F:\mathcal{X}\rightarrow \mathcal{Y}$ for $\mathcal{Y}_{u} \subset \mathcal{Y}$, where $\mathcal{Y}_{u}$ is the unseen set and $\{\mathcal{Y}_{i}\cap\mathcal{Y}_{u}=\emptyset, \forall i,j\}$.

\subsection{Mid-Level Semantic Knowledge Transfer}
\label{cha:fzsl_baseline}
To learn mid-level semantic knowledge transfer for federated learning, we formulate a baseline model which unifies federated learning and zero-shot learning in a decentralised learning paradigm.
Since generative based zero-shot learning is capable of generating pseudo image features according to a consistent and generic mid-level attribute space, in this work, we employ a representative ZSL method, i.e., f-CLSWGAN~\cite{xian2018feature}, as the backbone. %(in practice, our approach is compatible to various ZSL backbones, such as VAEGAN~\cite{xian2019f} and FREE~\cite{chen2021free}).
% %
% As for federated learning, we use the commonly used FedAvg~\cite{mcmahan2017communication}.
%
As shown in Fig.~\ref{fig:local_client}, the learning process consists of three iterative steps, namely local model learning, central model aggregation and local model reinitialisation with the central model.

In each local client, with the non-shared local data $\mathcal{S}_{i}{=}\left\{\bm{x}, y\right\}$, the model learning process follows f-CLSWGAN~\cite{xian2018feature}.
A generator $G(\bm{z}, \bm{a}_g)$ learns to generate a CNN feature \bm{$\hat{x}$} in the input feature space $\mathcal{X}$ from random noise $\bm{z}$ and a ground truth condition $\bm{a}_g$, where each value in $\bm{a}_g$ corresponds with one specific attribute, e.g. stripes. 
While a discriminator $D(\bm{x}, \bm{a}_g)$ takes a pair of input features $\bm{x}$ and a ground truth condition $\bm{a}_g$ as input and a real value as output.
Thus, the training objective of each local client model is defined as:

\begin{equation}
\label{eq:baseline_local_model}
\min _{G} \max _{D} \mathcal{L}_{W G A N}+\beta \mathcal{L}_{C L S},
\end{equation}
where $\beta$ is a hyper-parameter weight on the classifier, $\mathcal{L}_{W G A N}$ is wasserstein GAN loss~\cite{xian2018feature,gulrajani2017improved} and $\mathcal{L}_{C L S}$ is cross-entropy loss to classify generated CNN feature $\hat{x}$ to corresponding class label $y$.

After optimising each local client model for $E$ local epochs, the local model parameters \bm{$w_{i}$} are transmitted to a central server to aggregate a global model.
Following FedAvg~\cite{mcmahan2017communication}, the aggregating process is formulated as:
\begin{equation}
\label{eq:baseline_agg}  
    \bm{w}_{t}=\frac{1}{N \cdot S} \sum_{i \in N_{S}} \bm{w}_{i,t},
\end{equation}
where $N$ denotes the number of local clients and $t$ denotes the $t$-th global model iterative update round. $S$ denotes the randomly selected clients fraction for each round ($S\in[0.0,1.0]$) and $N_S$ is the set of selected clients.
Note that the central server only aggregates local model parameters without accessing local data so as to protect local data privacy.
Then, each local model is reinitialised with the central model as:
\begin{equation}
\label{eq:baseline_reinit}  
    \bm{w}_{i,t+1} =  \bm{w}_{t}.
\end{equation}
This is an iterative learning process (Eqs.(\ref{eq:baseline_local_model})-(\ref{eq:baseline_reinit})) until $T$ global model update round.
\sst{Since the attribute space is consistent among local clients, the learned global generator encodes mid-level semantic knowledge, i.e., given attribute vector $\bm{a}$ of a class, the generator can generate pseudo CNN features for this class regardless of whether it has been seen or not.}
% testing shitong 
Finally in the testing stage, based on the attributes from unseen classes, the learned generator from the global server is used to generate $M$ pseudo image features for each unseen classes $\mathcal{Y}_{un}$.
A softmax classifier is then trained under the supervision from pseudo features and tested for image classification on unseen classes.

\noindent\textbf{Selective Module Aggregation.}
Although aggregating holistic model parameters following FedAvg~\cite{mcmahan2017communication} is simple, it is inefficient for FZSL because the generic mid-level semantic knowledge is mainly encoded in the generator while the discriminator may contain knowledge specific to classes in each client. 
Inspired by recent approaches~\cite{wu2021collaborative,zhang2021fedzkt} in federated learning, we improve the baseline by decoupling the discriminator from the central model aggregation process, i.e., only aggregating the generator in the central server.
This not only reduces the cost for transmitting model parameters but also facilitates to learn more generalisable mid-level knowledge.
Thus, the central aggregation in Eq.~(\ref{eq:baseline_agg}) and the local client reinitialisation in Eq.~(\ref{eq:baseline_reinit}) are reformulated as: 
\begin{equation}
\label{eq:opt_agg}  
 \bm{w}_{G,t}=\frac{1}{N \cdot S} \sum_{i \in N_{S}}  \bm{w}_{G_{i},t},
\end{equation}
\begin{equation}
\label{eq:opt_reinit}  
 \bm{w}_{G_{i},t+1} =  \bm{w}_{G,t},~~~ \bm{w}_{D_{i},t+1} = \bm{w}_{D_{i},t},
\end{equation}
where $ \bm{w}_{G,t}$ and $ \bm{w}_{D,t}$ denote model parameters for a generator and a discriminator, respectively.

{\textbf{Discussion.}}
Note, we only aggregate the generator and decouple the discriminator in the selective module aggregation,
which slightly differs from the process in~\cite{mcmahan2017communication,wu2021collaborative,zhang2021fedzkt}.
Although selective module aggregation is not a significant contribution of this work,
our experimental results show that the improved strategy outperforms the baseline.
Thus, we use this useful strategy to improve a baseline model to facilitate the proposed mid-level semantic knowledge
transfer for FZSL.

%\subsection{Semantic Knowledge Augmentation for FZSL}
\subsection{\sgg{Semantic Knowledge Augmentation}}
Although the formulated baseline with selective module aggregation is able to transfer mid-level generic knowledge in a decentralized learning manner, it still suffers from sparse attribute and ambiguous attribute separability for limited data diversity in each client. 
To resolve this problem, we propose to explore a foundation model (language model (LM) RoBERTa~\cite{liu2019roberta} or text encoder of vision-language model (VLM) CLIP~\cite{radford2021learning} in this work)
 to explore semantic knowledge augmentation (SKA) to enrich the mid-level semantic space in FZSL.
Note we utilise off-the-shelf frozen text encoders directly which is disentangled from client-server aggregation process for efficient communication. 
Since a foundation model contains word embedding knowledge that can supply information regarding hierarchical relationships among classes, it can help FZSL to learn richer external knowledge with the sharable common attribute space.
%
% In this work, we introduce class-level and attribute-level semantic knowledge augmentation, where the former benefits more for coarse-grained inter-class separability while the latter is used to improve fine-grained inter-class separability.
%
In this work, we introduce class-level semantic knowledge augmentation, which greatly facilitates the generated feature diversification in both training and testing stages.
Empirically, we observe that directly concatenating a noise-enhanced text embedding and an attribute vector is an effective way, which do not require extra learnable parameters and can alleviate overfitting on seen classes.
%

% \noindent\textbf{{Class-Level Semantic Knowledge Augmentation.}
In our semantic knowledge augmentation, as shown in Fig.~\ref{fig:local_client}, we simply combine a default prompt `a photo of a' with class names and use this sentence as the input to the text encoder. 
We then further add the gaussian noise $ \bm{z}_c\sim N(0,\gamma)$ to the output text embedding $ \bm{a}_c$ so as to enrich the semantic space and to better align with the instance-wise diversified visual space, where each class-level semantic can always correspond to different samples with various poses and appearances in visual space.
The semantic augmented attribute is the concatenation between noise-enhanced text embedding and ground truth manual annotation attribute labels $ \bm{a}_g$. This semantic augmentation process is formulated as: 
\begin{equation}
\label{eq:SKA}  
 \bm{a}_{SKA} = [\bm{a}_c \oplus \bm{z}_c,\bm{a}_g ], 
\end{equation}
 where $\oplus$ is the element-wise summation.

During FZSL model training, the text embedding of seen class name is utilised as external knowledge to construct semantic knowledge augmented attribute $\bm{a}_{SKA}$ and further generate image features in each local client. 
The discriminator condition keeps \bm{$a_g$} to distinguish between the real distribution and the pseudo distribution.
% In test, the classnames of unseen classes are utilised for generating image features to train the classifier. 
In the testing stage, instead of generating pseudo image features based on the same attribute $\bm{a}_g$ for each class as in conventional ZSL~\cite{xian2018feature,xian2019f,chen2021free}, the SKA module supplies diversified attribute $\bm{a}_{SKA}$ for each class. 
The gaussian noise $\bm{z}_c$ in $\bm{a}_{SKA}$ can help explore the rich information in foundation model text encoder so to enrich the attribute space. 
Overall, our semantic knowledge augmentation can increase inter-class separability and supply diversified attribute space by only using text information of the class name.

\begin{table*}[t]
  \centering
  
  % \resizebox{\linewidth}{!}{%
   \begin{tabular}{l | l | c c c c c  }
      % & & \multicolumn{5}{c}{\textbf{FZSL}}  \\
      \hline
      & Method & \textbf{AWA2} & \textbf{AWA1} & \textbf{aPY} &  \textbf{CUB} & \textbf{SUN}  \\
      \hline
      \multirow{3}{*}{$Centralised$}
      % comment the reported result
      %  &CLSWGAN~~\cite{xian2018feature} & $ -  $  & $68.2 $ &$ - $ &  $57.3$ & $60.8 $ \\ 
       
       &VAEGAN~~\cite{xian2019f} & $ 61.4 $  & $55.9 $ &$ 18.6 $ &  $ 44.8$ & $ 56.9 $ \\ 
       &CLSWGAN~~\cite{xian2018feature} & $ 67.4$  & $66.6 $ &$37.7 $ &  $56.8$ & $60.3 $ \\  
       &FREE~~\cite{chen2021free} & $ 67.7 $  & $68.9 $ &$ 42.2 $ &  $60.9 $ & $61.3 $ \\

       \hline
      \multirow{7}{*}{$Decentralised$}
      &CLSWGAN+FedAvg~\cite{mcmahan2017communication} & $61.6$ & $58.5$ & $ 33.8$  & $53.8$& $ 59.5$ \\
      &CLSWGAN+FedProx~~\cite{li2020federated}  & $ 61.3 $ & $ 58.4 $ & $ 34.0$ & $ 53.1 $ & $59.3 $ \\
      &CLSWGAN+MOON~~\cite{li2021model}   & $ 61.0$ & $ 58.6$ & $ 33.2 $ & $\underline{55.1} $ & $59.5 $ \\

      % &FL-VAEGAN  & $48.9 $ & $44.0 $ & $16.4 $ & $43.6 $ & $56.2 $ \\ Generator Discriminator size not the same as paper report
      &FL-VAEGAN~\cite{xian2019f}  & $60.2 $ & $51.6 $ & $18.7 $ & $43.2 $ & $55.5 $ \\  %Generator Discriminator size 4096
  
      % &FL-VAEGAN+SMA  & \underline{$50.4$} & \underline{$44.6 $} & $\textbf{25.9} $ & \underline{$46.0 $} & \underline{$59.4 $} \\
      % &FL-VAEGAN+SMA+SKA(VLP)  & $\textbf{60.1} $ & $\textbf{58.2} $ & \underline{$ 19.6$} & $\textbf{52.6} $ & $\textbf{61.2} $ \\
      % &FL-VAEGAN+SMA+SKA(LM)  & $ $ & $ $ &  &  &  \\
  
      % \cdashline{2-7}
      &FL-FREE~\cite{chen2021free}  & $60.9$  & $59.8 $& $ 25.9$  & $ 54.5$ & $56.4 $\\
      % &FL-FREE+SMA  & \underline{$61.4$} & \underline{$61.1$} & \underline{$27.4$} & \underline{$55.4$} & \underline{$57.0$} \\
      % &FL-FREE+SMA+SKA(VLP)  & $\textbf{ 68.4}  $ & $\textbf{68.4} $ & $\textbf{32.0} $ & $\textbf{60.7} $ & $\textbf{60.5} $ \\
      % &FL-FREE+SMA+SKA(LM)  &  &  &  &  &  \\
  
      % \cdashline{2-7}
      \cdashline{2-7}
      % &FL-CLSWGAN+SMA  & \underline{$62.8$} & \underline{$61.7$} & \underline{$ 38.4$ } & \underline{$55.5$} & \underline{$ 59.4$}\\
      &Ours (LM)  & \underline{65.4}  & \underline{64.6} & \underline{41.6} & 54.8 &  \underline{61.1} \\
      &Ours (VLM)   & $\textbf{69.0}$& $ \textbf{70.6}$ & $\textbf{47.1}$  & $\textbf{59.4} $ & $ \textbf{66.5}$\\

      \hline
  \end{tabular}
  % }
  \caption{Comparison with the related methods on AWA2, AWA1, aPY, CUB and SUN (Top-1 accuracy).
  SKA(LM) and SKA(VLM) denote semantic knowledge augmentation with text encoder from language model and vision-language model respectively. \textbf{Bold} and \underline{underline} are the best and the second best results.
  Note that centralised methods are used for reference only because they are not direct counterparts.
  } 
  \label{table:FZSL}
  \end{table*}

{\textbf{Discussion.}}
In this work, we consider a text encoder of a foundation model as a generic off-the-shelf model trained with a vast quantity of data.
When using a text encoder of a vision-language foundation model (e.g., CLIP~\cite{radford2021learning}),
the text encoder has been trained with both texts and images so the model is exposed to plausibly similar images in training with the potential for zero-shot learning.
However, since we only use the text encoder (not the whole vision-language model as PromptFL~\cite{guo2022promptfl})
and the foundation model is defined as a generic off-the-shelf model,
the text encoder can be used in our approach to explore semantic knowledge augmentation to enrich mid-level semantic space in FZSL.
Besides, our approach can also be configured with a language model (e.g., RoBERTa~\cite{liu2019roberta}),
which has not been trained with any images, to explore semantic knowledge augmentation for FZSL.
Overall, with the improved baseline and semantic knowledge augmentation,
our approach is capable of learning mid-level knowledge to facilitate FZSL.
Since training data are non-shared and mid-level knowledge only contains semantic properties insensitive to objects of interests, our approach inherently protects privacy.

\section{Experiments}
\textbf{Datasets.} To evaluate the effectiveness of our approach, we conduct extensive experiments on five zero-shot benchmark datasets,
including three coarse-grained datasets: (Animals with Attributes (AWA1)~\cite{lampert2013attribute}, Animals with Attributes 2 (AWA2)~\cite{xian2018zero} and Attribute Pascal and Yahoo (aPY)~\cite{farhadi2009describing}); 
and two fine-grained datasets (Caltech-UCSD-Birds 200-2011 (CUB)~\cite{wah2011caltech} and SUN Attribute (SUN)~\cite{patterson2012sun}).
AWA1 is a coarse-grained dataset with 30475 images, 50 classes and 85 attributes, while AWA2 shares the same number of classes and attributes as AWA1 but with 37322 images in total.
The aPY dataset is a relatively small coarse-grained dataset with 15339 images, 32 classes and 64 attributes.
CUB contains 11788 images from 200 different types of birds annotated with 312 attributes, while SUN contains 14340 images from 717 scenes annotated with 102 attributes.
We use the zero-shot splits proposed by~\cite{xian2018zero} for AWA1, AWA2, aPY, CUB and SUN ensuring that none of training classes are present in ImageNet~\cite{russakovsky2015imagenet}.
All these five datasets are composed of seen classes set and unseen classes set.
In decentralised learning experiments, we evenly split the seen classes set randomly to four clients.
Note, both seen classes and unseen classes share the same attribute space in each dataset.

\noindent\textbf{Evaluation Metrics.}
In FZSL, the goal is to learn a generalisable server model which can assign unseen class label $\mathcal{Y}_{u}$ to test images.
Following commonly used zero-shot learning evaluation protocol~\cite{xian2018zero}, the accuracy of each unseen class is calculated independently before divided by the total unseen class number, i.e., calculating the average per-class top-1 accuracy of the unseen classes. 
%

% \begin{figure}[t]
%   \centering
%   \subfigure[ss]{
%   \begin{minipage}
%     \includegraphics[width=0.5\textwidth]{./image/client_img_distribution_larger.png}  
%   \end{minipage}
%       }

%   \subfigure[ss]{
%   \begin{minipage}
%     \includegraphics[width=0.5\textwidth]{./image/client_img_distribution_larger.png}  
%   \end{minipage}
%   }

  % \caption{
  %  %\cite{zhu2019deep}
  %  Detailed class-level partitions on AWA2. The depth of color represents the number of samples in each class.
  %  %
  %  % Mid-level transfer can only leak class label with the help of attribute-label map, while high-level transfer leak the raw images.
  % }
  % \label{fig:client_distribution}
% \end{figure}

\begin{figure}[htbp]
  \centering
  \begin{minipage}[t]{0.49\textwidth}
    \centering
    \includegraphics[width=\textwidth]{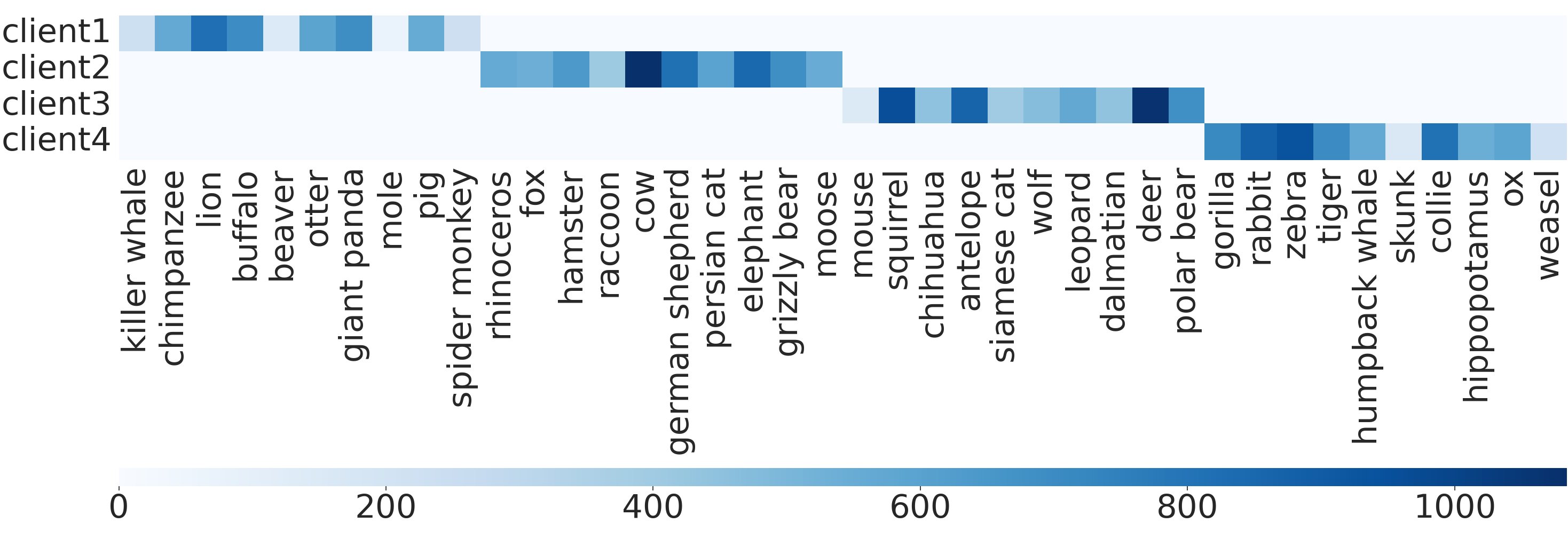}
    (a) High-level (class) partition for each client on AWA2.
    % \subcaption{ Detailed high-level (class) partitions for each client on AWA2. The depth of color represents the number of samples in each class.}
  \end{minipage}
  % \hfill
  \begin{minipage}[t]{0.49\textwidth}
    \centering
    \includegraphics[width=\textwidth]{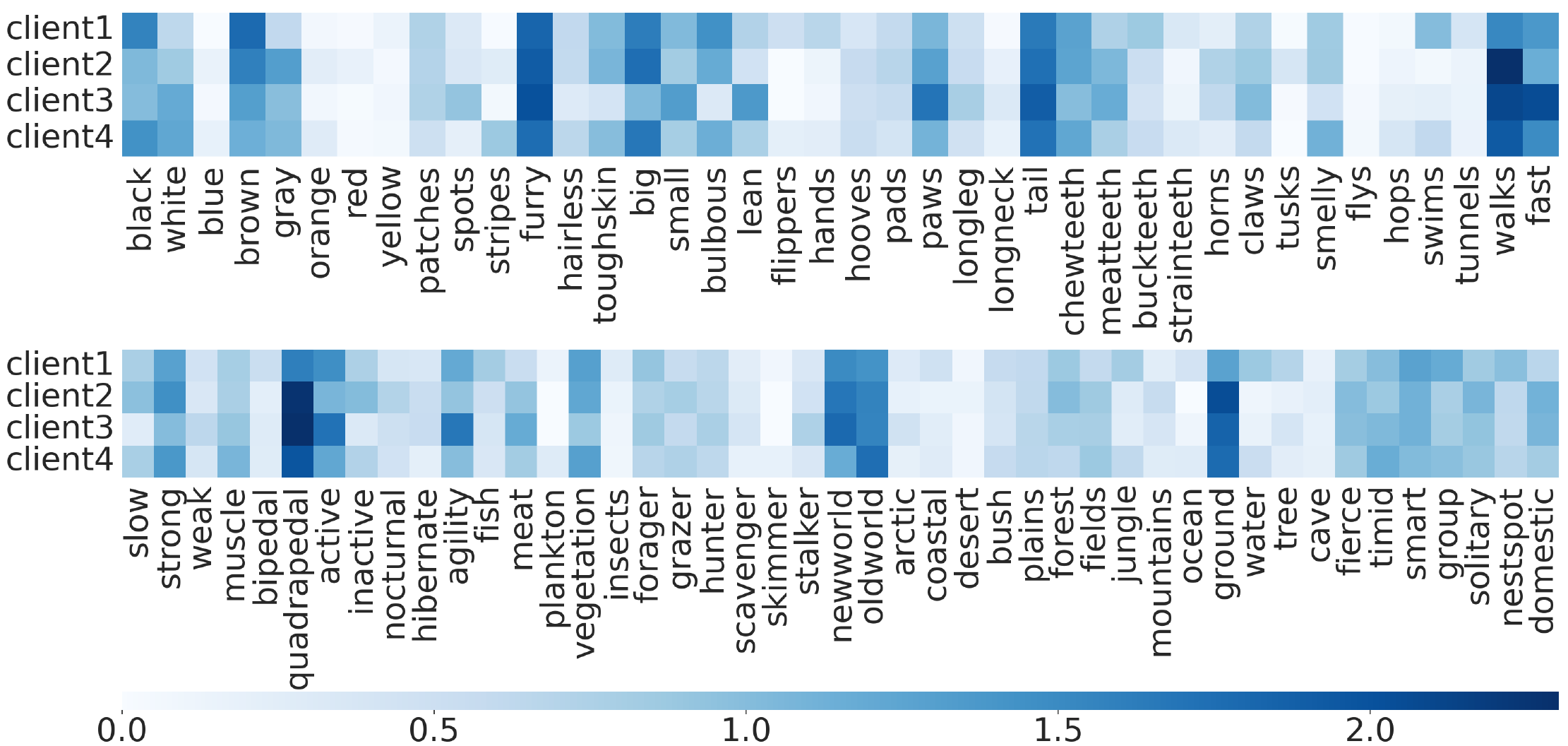}
    % \caption{Detailed mid-level (attribute) partitions for each client on AWA2. The depth of value represents the sum of each continues attributes~\cite{xian2018zero} for all classes in each client. 
    % }
    % \label{fig:feat_distribute}
    (b) Mid-level (attribute) partition for each client on AWA2.
  \end{minipage}
  \caption{ High-level vs mid-level partition for each client on AWA2. The depth of color represents the number of samples for each class in (a) and the sum of each continues attributes~\cite{xian2018zero} for all classes in each client in (b).}
  \label{fig:cls_distribute}

\end{figure}

% \begin{figure}[t]
%   \centering

%   \includegraphics[width=0.5\textwidth]{./image/confusion_feature_c.png}
%   \caption{
%    %\cite{zhu2019deep}
%    Detailed attribute-level partitions on AWA2. The depth of value represents the sum of continues attributes~\cite{xian2018zero} for all classes in each client. 
%    %
%    % Mid-level transfer can only leak class label with the help of attribute-label map, while high-level transfer leak the raw images.
%   }
%   \label{fig:client_distribution_feature}
% \end{figure}

\begin{figure}[t]
    \centering
    \includegraphics[width=0.5\textwidth]{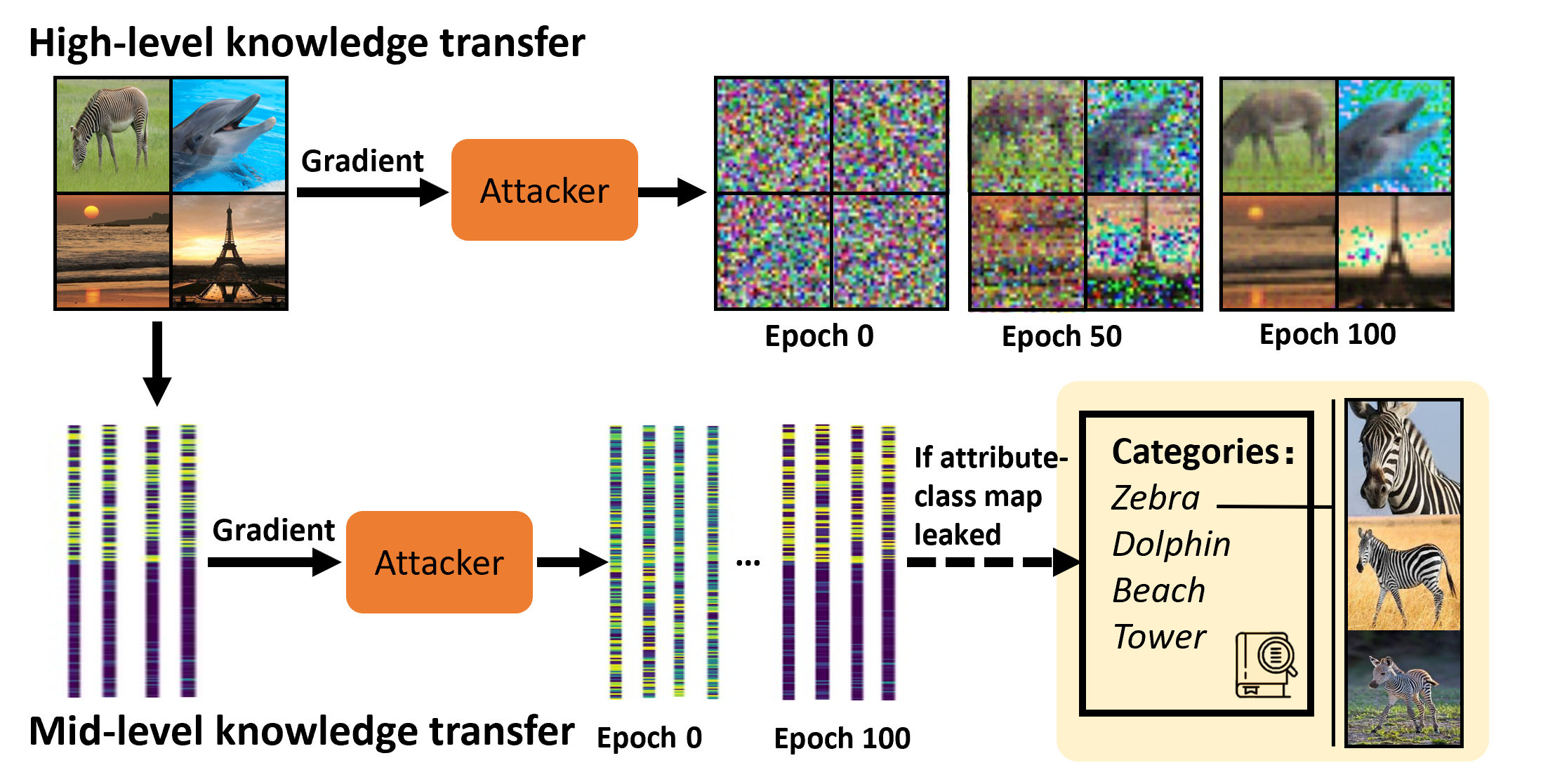}
    \caption{
     %\cite{zhu2019deep}
     The gradient leakage attack on high-level and mid-level knowledge transfer. 
     %
     % Mid-level transfer can only leak class label with the help of attribute-label map, while high-level transfer leak the raw images.
    }
    \label{fig:attack}
  \end{figure}

  \noindent\textbf{Implementation Details.}
  Following~\cite{xian2018feature,xian2019f,chen2021free}, in our approach, we employed a frozen ResNet-101~\cite{he2016deep} pretrained on ImageNet~\cite{russakovsky2015imagenet} as the CNN feature extractor and constructed our baseline model with a  generator and a discriminator for each client respectively following the representative zero-shot learning work~\cite{xian2018feature}. 
  Both generator and discriminator have a simple architecture consisting of only two MLP layers.
  Further, we employed a frozen pretrained text encoder (base RoBERTa~\cite{liu2019roberta} in SKA (LM) and text encoder of ViT-Base/16 CLIP~\cite{radford2021learning} in SKA(VLM)) to supply class-name-based text embedding for each client.
  All clients share the same model structure while the server aggregates local model parameters to construct a global model. 
  %
  % With selective module aggregation (SMA), only the generators from local clients are aggregated.
  As for further improvement with semantic knowledge augmentation (SKA), the frozen text encoder is kept locally and not aggregated to the server. 
  % both the generator and text-enhanced module are aggregated to the server. 
  %
  % We employed class-level semantic knowledge augmentation for three coarse-grained datasets (AWA1, AWA2 and aPY) while both class-level and attribute-level augmentation were used for two fine-grained datasets (CUB and SUN).
  % 
  % Each client contains local non-overlapping classes from the seen classes set and the aggregated server model is tested on the unseen classes set.
  %
  By default, we set the number of local clients $N$=4, random client selection fraction $S$=1 and noise augmentation $\gamma$ to 0.1.
  Generated feature number $M$ and classifier weight $\beta$ follows~\cite{xian2018feature}.
  We empirically set batch size to 64, maximum global iterations rounds $T$=100, maximum local epochs $E$=1.
  For each local client, we used Adam optimizer with a learning rate of $1e{-}3$ for CUB, $2e{-}4$ for SUN and $1e{-}5$ for the others.
  Our models were implemented with Python(3.6) and PyTorch(1.7) and trained on NVIDIA A100 GPUs.

  \subsection{Federated Zero-Shot Learning Analysis}
  There are no existing works discussing mid-level semantic knowledge transfer in federated learning,
  so besides our baseline model (CLSWGAN~\cite{xian2018feature} with FedAvg~\cite{mcmahan2017communication}),
  we implement (1) a traditional ZSL method VAEGAN~\cite{xian2019f} and a recent ZSL method FREE~\cite{chen2021free} with FedAvg denoted as FL-VAEGAN and FL-FREE respectively,
  and (2) CLSWGAN with two other federated learning paragidms, namely FedProx~\cite{li2020federated} and MOON~\cite{li2021model}.
  % %
  % Further, the proposed SMA and SKA are implemented on the baseline FL-CLSWGAN, with pretrained language model based SKA (LM)  
  % and text encoder of pretrained vision-language model based SKA (VLM) respectively. % where the generality and compatibility of SMA and SKA can be demonstrated. 
  %
  % Note, when implementing SMA to FREE, feature refinement module will also be aggregated to the server which will be used during testing. 
  % 
  All compared methods are inductive where only attribute information of unseen classes are used for training the classifier and unseen images are not used during training.
  % TODO: add to related work

  From Table~\ref{table:FZSL}, we can see that:
  (1) Compared with the centralised method (CLSWGAN), the decentralised baseline (CLSWGAN+FedAvg) yields compelling performance,
  showing the effectiveness of the proposed paradigm for learning globally generalised model whilst protecting local data privacy.
  \sst{This can be attributed to the learnt generator which can
    effectively extract mid-level semantic knowledge \sgg{in either} decentralised settings with \sgg{disjoint} class label spaces \sgg{or} centralised settings.
  This observation brings inspiration to federated learning with
  non-IID data and proves that semantic mid-level information
  (\sgg{attributes in particular}) can be the consistent and generalisable knowledge extracted from heterogeneous clients so to boost all the participants.}
  %
  % (2) With selective module selection (SMA), overall the performance of the baselines are improved (3.4\% in FL-VAEGAN, 1\% in FL-FREE and 2.1 \% in FL-CLSWGAN on average), 
  % (2) With selective module selection (SMA), overall the performance of the baselines are improved 2.1 \% in FL-CLSWGAN on average,
  % which verifies that learning a generic generator and decoupling the discriminator from central aggregation can facilitate mid-level semantic knowledge transfer in FZSL;
  %
  (2) Our approach significantly improves the baseline by 9\% with SKA(VLM) and by 4\% with SKA(LM) on average, which validates the effectiveness of our approach for FZSL.
  \sst{This can be attributed to the text encoder of foundation models which can supply rich semantic space and increase the diversity of generated pseudo samples to a large degree, which can further promote the generalisability of the model.}
  (3) Comparing with other FL approaches (FedProx and MOON) and ZSL approaches (FL-VAEGAN and FL-FREE), our approach achieves significantly better performance on average, showing the adaptation of our method in learning mid-level semantic knowledge in federated learning.
  %
  % \sst{Previous works are not designed for federated zero shot learning and not contribute to federated setting with nonoverlap class distribution (for VAEGAN, FREE), yet no benefit to unseen class inference(for FedProx, MOON).}

  \subsection{High-Level vs. Mid-Level Knowledge Transfer}
  \textbf{Privacy Protection.}
  To verify the effectiveness of privacy protection from transferring
  mid-level knowledge \sgg{by attributes}, we performed the gradient
  leakage attack~\cite{zhu2019deep} 
  on federated learning with high-level knowledge transfer and our mid-level knowledge transfer. 
  As shown in Fig.~\ref{fig:attack}, the source images used in high-level knowledge transfer are recovered by the attack, which cause privacy leakage. 
  In contrast, in our mid-level knowledge transfer, only attribute vector can be recovered,
  \sst{which is a continuous embedding~\cite{xian2018zero} in our case and only leak less sensitive information.
  In extreme cases, even if the attribute-label mapping is further leaked to the attacker, the attacker can only know the corresponding categories (class-level) with millions of possibilities but not the exact object raw images (instance-level).
  This demonstrates high privacy security and robustness against attacks in our paradigm.
  
  \textbf{Model Generalisation.}
  To supply a qualitatively analysis of model generalisation from
  transferring mid-level knowledge \sgg{by attributes}, we visualise
  data partitions for FZSL on both high-level (class) and mid-level
  (attribute).
  In the perspective of high-level (class) partition as shown in Fig.~\ref{fig:cls_distribute} (a), each client contains non-overlapping class label space which results to high degree of class label distribution skewness. 
  %  and has a notable performance decrease (76.25\% on ResNet50 with CWT~\cite{chang2018distributed} when testing on seen classes) comparing with IID data partition reported by Qu et al.~\cite{qu2022rethinking}.
  % %
  Qu et al.~\cite{qu2022rethinking} reported a notable performance decrease(76.25\% on ResNet50) when comparing this extremely heterogeneous partition with IID data partition.
  %
  % Even harder in our case, we test the model on unseen classes, which makes it less possible by transferring only high-level knowledge. 
  %
  However, in the perspective of mid-level (attribute) partition as shown in Fig.~\ref{fig:cls_distribute} (b), it shows an overlapping attribute space and results to a lower degree of attribute label distribution skewness.
  %
  % It is observed that different clients share less heterogeneous underlying mid-level attribute knowledge, which makes it possible to generalise to unseen classes. 
  %
  Experiment results on Table.~\ref{table:FZSL} and Table.~\ref{table:local_decentralised} prove the generalisability of transferring mid-level knowledge to unseen classes. 
  }

\subsection{Local Training vs. Decentralised Learning}
To verify the effectiveness of the formulated federated zero-shot learning paradigm, we separately train four individual %baseline
local models with local client data and compare with decentralised learning models.
\sst{For a clearer comparison, we further test the model in centralised learning with and without our proposed SKA.}
The performance are tested on the same unseen classes for all compared methods.
As shown in Table \ref{table:local_decentralised}, the decentralised baseline significantly outperforms all individual client models.
This shows that the federated collaboration between the localised clients and the central server model facilitates to optimise a generalisable model in FZSL.
\sst{Moreover, the shared mid-level knowledge among clients can be extracted from non-overlapping isolated local data with privacy protection and help boost the discriminative ability for unseen classes.
Additionally, we evaluate SKA (VLM) with centralised method, which also shows a significant improvement and proves the effectiveness of proposed SKA in both centralised and decentralised scenario.}

\begin{table}[t]
    \begin{center}
    
    \begin{tabular}{c|c|c|c|c|c|c}
    \hline
     SMA &  SKA & AWA2 & AWA1 & aPY &CUB & SUN \\
    
    \hline\hline
  
    \ding{55} & \ding{55} &  61.6 &  58.5 & 33.8 & 53.8  &59.5\\   
  
    \ding{51} &  \ding{55} & 62.8 &  61.7 & 38.4 & 55.5  &59.4\\   
    \hdashline
    \ding{55}&  \ding{51}(LM) & 65.0  & 64.2 & 39.1 & 54.1  & 60.6  \\
    \ding{51}&  \ding{51}(LM)  & 65.4 & 64.6  & 41.6 & 54.8  & 61.1  \\
    % \hdashline
    % \ding{55}&  \ding{51}(VLM)& 69.3 & 70.0 & 42.9 & 58.5  & 65.6  \\
    \ding{51}&  \ding{51}(VLM) & 69.0 & 70.6  & 47.1 &  59.4 & 66.5  \\
    
    \hline
    \end{tabular}
    \end{center}
    \caption{Baseline with proposed module variations.
    SMA: selective module aggregation.
    SKA: semantic augmentation with RoBERTa as language model (LM) and CLIP text encoder as vision-language model (VLM) respectively.}
    \label{table:ska}
  \end{table}

  \subsection{Ablation Study}
\begin{figure}[t]
  \centering
  \includegraphics[width=0.48\textwidth]{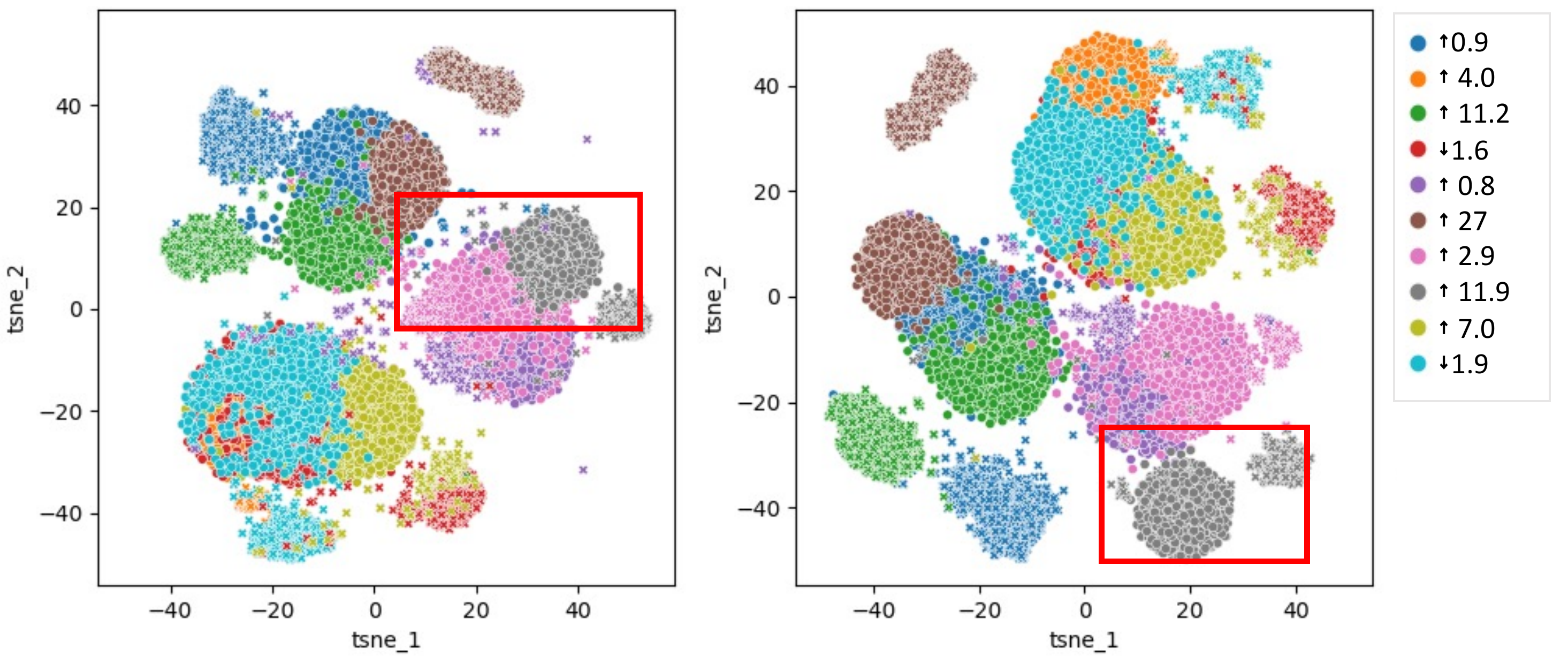}
  \caption{tSNE of unseen classes on AWA2 for baseline+SMA (left) and baseline+SMA+SKA (VLM) (right).
  The same colour implies the same class.
  Circle and cross means the generated distribution and real unseen distribution, respectively.
  The number in the caption means increase or decrease percentage for each class after implementing SKA (VLM). 
  The classifier trained on generated pseudo distribution is tested on the unseen real distribution.
  }
  \label{fig:tsne}
  \end{figure}

  % \paragraph{Component Effectiveness Evaluation.}
  % \textbf{Component Effectiveness Evaluation.}
  \textbf{Component Effectiveness Evaluation.}
  % \paragraph{SMA SKA}
  % A large scale pretrained model can usually supply rich information and may benefit decentralised zero shot learning.
  %
  % As shown in Table~\ref{table:FZSL}, the performance of the baseline model can be significantly improved with selective module aggregation (SMA) and semantic knowledge augmentation (SKA).
  %
  To evaluate the component of our approaches, We further analyse the results both quantitatively and qualitatively.
  Quantitatively, 
  we report experimental results in Table~\ref{table:ska} 
  for the baseline with and without SMA, SKA (LM) and SKA (VLM) respectively.
  It can be observed from Table~\ref{table:ska} that SMA can bring benefits (except baseline on SUN) with and without SKA, which can 
  demonstrate the generality and effectiveness of the SMA module to build an improved baseline for FZSL.
  Besides, SKA can bring significant improvement in all of the five datasets (except SKA(LM) in CUB), proving that the diversity supplied by SKA can help the global model generalisability. 
  Qualitatively, the tSNE visualisations of AWA2 unseen classes for baseline+SMA before and after implementing the semantic knowledge augmentation with vision-language model are shown in Fig.~\ref{fig:tsne}.
  It can be seen that with SKA (VLM), the generated distribution has a larger inter-class distance as shown in the red box. 
  This larger inter-class distance improves coarse-grained classification accuracy.
  % , which is consistent with the conclusion of FREE~\cite{chen2021free}.

  \begin{table}[t]
    \centering
    
    \begin{tabular}{c|c|c|c|c|c|c }
    \hline
    % \multicolumn{1}{c|}{\multirow{1}{*}{Text}} & \multicolumn{1}{c|}{\multirow{1}{*}{Methods}}
        \multicolumn{2}{c|}{Text Encoder}  %& \multicolumn{1}{c|}{\multirow{1}{*}{Methods}}
  
    & AWA2 & AWA1 & aPY & CUB & SUN \\
    \hline\hline
    %   \ding{55} &\ding{55} &62.8 & 61.7& 38.4& 55.5 & 59.4\\
       \multicolumn{2}{c|}{\ding{55}}  &62.8 & 61.7& 38.4& 55.5 & 59.4\\
  
        \hline
    \multicolumn{1}{c|}{\multirow{2}{*}{\tabincell{c}{LM}}}
    % &CLSWGAN & 68.5 & 68.2 &  57.3 & 60.8 \\
     &BERT & 63.4 & 63.8 & 41.1 & 54.6 & 60.9 \\    
    &RoBERTa & 65.4 & 64.6 & 41.6 & 54.8 & 61.1 \\
    % &SBERT & 44.1 &  & &\\
    % &GloVe & 53.7 &  & &\\
    \hline
    \multicolumn{1}{c|}{\multirow{2}{*}{\tabincell{c}{VLM}}}
    &DeFILIP  & 74.1 & 75.5 & 49.4 & 58.2 & 64.2  \\
    &CLIP  & 69.0 & 70.6 & 47.1 & 59.4 & 66.5  \\
    \hline
  
    \end{tabular}
    \caption{In comparison with Baseline+SMA, evaluation with the text embedding of two pretrained Language Models (LM) and two pretrained Vision-Language Models (VLM) are reported.}
    \label{table:exp_textEmb}
    \end{table}

\begin{figure*}[t]
    \centering
    \subfigure[Backbone Variations]{
      \includegraphics[width=0.245\linewidth]{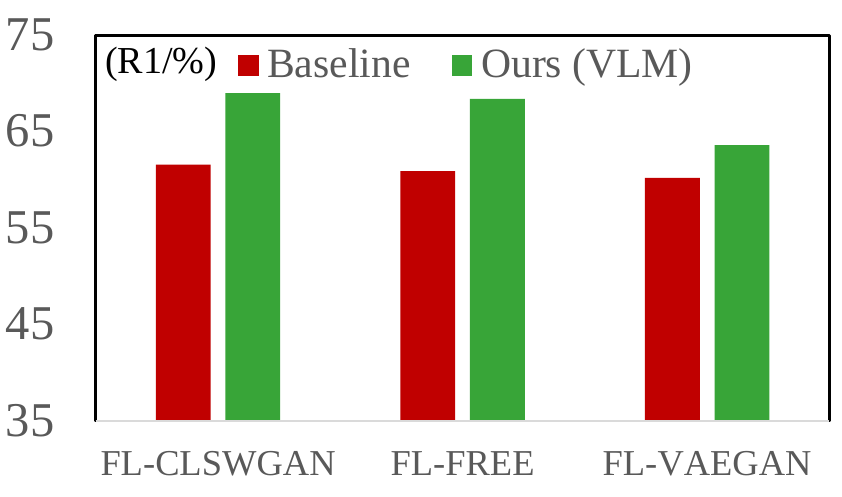}
      \label{fig:backbone_variation}
      }
    \subfigure[Client Number]{
    \includegraphics[width=0.22\linewidth]{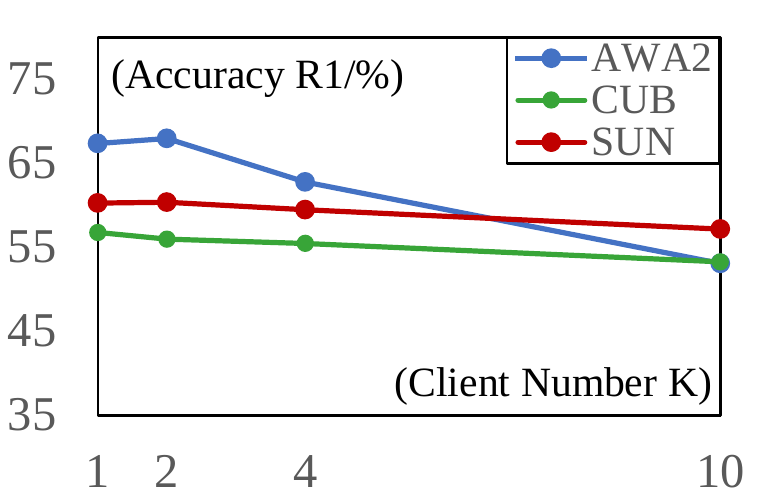}
    \label{fig:client_num}
    }
    \subfigure[Client Fraction]{
    \includegraphics[width=0.225\linewidth]{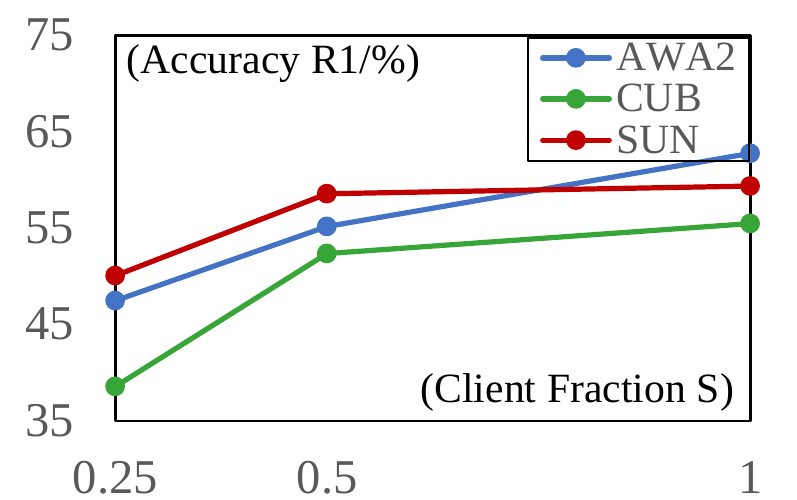}
    \label{fig:client_frac}
    }
    \subfigure[Client Local Step]{
    \includegraphics[width=0.213\linewidth]{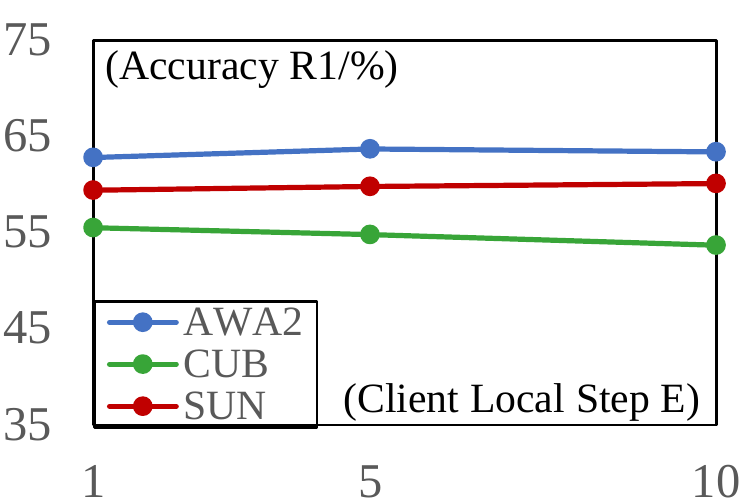}
    \label{fig:local_step}
    }

    \caption{Ablation study on (a) client number, (b) client fraction,
    (c) local steps, (d) backbone variations.}
    \end{figure*}

  \textbf{Variation of Semantic Knowledge Augmentation.}
  In Table~\ref{table:exp_textEmb}, we study more variants of text encoder in semantic knowledge augmentation.
  Here, BERT~\cite{devlin2018bert} is a bidirectional encoder similar to RoBERTa, while DeFILIP~\cite{cui2022democratizing} is a variation of CLIP.
  % %
  % DeFILIP is a variation of CLIP~\cite{radford2021learning} which aims to explore fine-grained information in a more data efficient method. 
  % FZSL can gain benefit from a large scale pretrained text encoder.
  % %
  % We naturally interested in whether other language models or visual language pretrained models can bring benefits. 
  % %
  % We therefore compare one large scale language models BERT~\cite{devlin2018bert}; and the text encoder of a vision-language pretrained model DeFILIP~\cite{cui2022democratizing}.
  % %
  % BERT, similar to RoBERTa, is a bidirectional encoder trained on 16GB text corpora. 
  % %
  % DeFILIP is a variation of CLIP~\cite{radford2021learning} which aims to explore fine-grained information in a more data efficient method. 
  % %
  % Glove has no an explicit encoder and returns the average embedding of each vocabulary of the input sentence, which is composed of a manual designed prompt `a photo of a ' and the classname.
  %
  % Both BERT and DeFILIP will calculate the embedding of the whole input sentence, where we fed in the same sentence as our SKA (LM) and SKA (VLM). 
  %
  As shown in Table \ref{table:exp_textEmb}, we can see that:
  (1) Both LM and VLM text encoder can bring benefits comparing with baseline, which shows the effectiveness and generality of the proposed SKA structure.
  % (1) Both LM and VLP text encoder can bring benefits (except LM model cannot guide fine-grained feature properly) comparing with baseline, which can demonstrate the effectiveness and generality of the proposed SKA structure.
  %
  (2) Our approach with VLM achieves better results compared to ours with LM. 
  The reason is mainly that the text encoders in VLM are pretrained with both texts and images so it is easier for these text encoders to
  achieve the alignment between visual and semantic distribution in FZSL.
  (3) DeFILIP, a fine-grained variation of CLIP, achieves the best result among different text encoders. 
  Interestingly, we find that DeFILIP with attribute-level SKA (text
  input: `a photo of  a \{attribute\} \{class name\}.') can achieve
  59.8\% and 65.6\% on CUB and SUN respectively (cf. 58.2\% and 64.2\%
  on CUB and SUN with class-level SKA), which implies that the
  fine-grained information from DeFILIP can be further explored
  \sgg{by feature mining}.

  \begin{table}[t]
    \centering
    
    \begin{tabular}{c|c|c|c|c|c|c}
    \hline
    \multicolumn{1}{c|}{\multirow{1}{*}{Setting}} & \multicolumn{1}{c|}{\multirow{1}{*}{Methods}}
    & AWA2 & AWA1 & aPY & CUB & SUN \\
    
    \hline\hline
    \multicolumn{1}{c|}{\multirow{5}{*}{\tabincell{c}{Local}}}
    &Client 1 & 49.0 & 47.8 & 23.2 & 42.4 & 50.6 \\
    
    &Client 2 & 37.1 & 38.7 & 22.8 & 40.5 & 52.1   \\
    
    &Client 3 & 40.2 & 41.1 & 34.3 & 40.2 &  49.8\\
    
    &Client 4 & 53.0 & 51.9 & 26.3 & 40.2 & 50.4\\
    \cline{2-7} 
    
    & Average & 44.8 & 44.9 & 26.7 & 35.5 & 50.7\\
    
    \hline
    %   \multicolumn{1}{c|}{\multirow{2}{*}{\tabincell{c}{Ensembles}}} & Feat-Concatenation &  &  \\
    
    %   & Parameter-Average &  &  \\
    %   \hline
    \multicolumn{1}{c|}{\multirow{2}{*}{\tabincell{c}{Decen-\\tral}}} 
    & Baseline & {61.6 }& {58.5 }&{33.8} & {53.8 }& {59.5 }\\
    & Ours & 69.0 & 70.6 & 47.1 & 59.4 & 66.5 \\
    \hline
    \multicolumn{2}{c|}{Centralised} & 67.4 & 66.6 & 37.7  & 56.8 & 60.3\\
    \hline
    \multicolumn{2}{c|}{Centralised+SKA(VLM)} & 72.8 & 70.1 & 46.5  & 58.0 & 64.9\\
    \hline
    \end{tabular}
    \caption{Comparing local training and decentralised learning.
    % Top-1 accuracy in percentage on unseen classes. 
    Baseline denotes CLSWGAN with FedAvg, while ours denotes Baseline+SMA+SKA(VLM).
    }
    \label{table:local_decentralised}
    \end{table}

 \begin{table}[t]
    \centering
    
    \begin{tabular}{c|c|c|c|c}
    \hline
    % \multicolumn{1}{c|}{\multirow{1}{*}{}} & \multicolumn{1}{c|}{\multirow{1}{*}{GT}}
     Features & Methods &AWA2  &CUB & SUN \\
    
    \hline\hline
  
    % \multicolumn{}{}{\multirow{3}{*}{\tabincell{c}{}}} 
    & Baseline & 55.4 & 45.4  & 51.1  \\ 
    GoogLeNet & Ours(LM) &  \underline{60.6}  & \underline{46.3}  & \underline{52.4}  \\ 
    & Ours(VLM) & \textbf{65.2}  & \textbf{50.9} &  \textbf{55.8} \\   
    \hdashline
    &Baseline & 63.0   & 56.5   & 57.9  \\ 
    MobileNetV2 &Ours(LM)& \underline{66.3} &   \underline{57.1}  & \underline{58.1}  \\ 
    &Ours(VLM) & \textbf{70.9} & \textbf{61.5}  & \textbf{63.8} \\ 
    \hdashline
    &Baseline & 61.6 & 53.8 &  59.5 \\
    ResNet101 & Ours(LM)& \underline{65.4}  &  \underline{54.8} &  \underline{61.1} \\
    &Ours(VLM) & $\textbf{69.0}$  & $\textbf{59.4} $ & $ \textbf{66.5}$\\
  
    \hline
    \end{tabular}
    \caption{Evaluation with different CNN feature encoders.
    }
    \label{table:smaller}
  \end{table}
  
\textbf{Variation of Encoder Architectures.}
In this work, following~\cite{xian2018feature,xian2019f,chen2021free}, we mainly use ResNet101 as the CNN feature encoders.
In Table~\ref{table:smaller}, we further report the results of our approach with smaller CNN feature encoders, including GoogLeNet~\cite{szegedy2015going} and MobileNetV2~\cite{sandler2018mobilenetv2}.
From Table~\ref{table:smaller}, we can observe that with CNN features extracted from different backbones,
our approach can consistently improve the performance of the baseline. 
%
% Specifically, with SMA and language model based SKA, the average improvement is 2.5\% in GoogLeNet features and 2.1\% in ResNet101 features. With SMA and text encoder of vision-language model based SKA, the average improvement is 6.7\% in both GoogLeNet and ResNet-101 features.
%.
This shows that \sgg{our method is not limited to} ResNet-101
features. \sgg{It is} also applicable to other more compact and smaller network features.
%extracted CNN features from more compact and smaller network.% models.

\begin{table}[t]
  \begin{center}
  
  \begin{tabular}{l | c|c|c|c|c|c}
  \hline
  &Methods & AWA2 & AWA1 & aPY &CUB & SUN \\
  
  \hline\hline
  \multirow{2}{*}{$Evenly$}

  & FL-CLSWGAN &  61.6 &  58.5 & 33.8 & 53.8  &59.5\\   

  & Ours (VLM) & 69.0& 70.6 & 47.1  & 59.4  & 66.5\\  
  \hline
  \multirow{2}{*}{$Unevenly$}

  &FL-CLSWGAN &  63.0 &  58.7 & 35.3 & 55.8  &59.7\\   

  &Ours (VLM) & 69.3 &  70.3 & 46.2 & 59.4  &66.1\\   
    
  \hline
  \end{tabular}
  \end{center}
  \caption{Comparing baseline and ours (VLM) with both evenly and random unevenly split. FL-CLSWGAN donates CLSWGAN+FedAvg in Table~\ref{table:FZSL}.}
  \label{table:uneven}
\end{table}

\sst{
\textbf{Clients with Uneven Split.}
\sgg{We further} discuss the influence of data distribution with even or uneven class label split. 
In this work, we mainly conduct \sgg{experiments} with the evenly
split client data distribution, where each client has equally number
of non-overlapping class labels. 
Although even split of class labels for each client is already the extremely non-IID data partition with high degree of label skewness~\cite{qu2022rethinking}, we further test on uneven split non-overlapping class labels. 
Specifically, the seen class labels are randomly split to 4 clients and ensure that each client has at least one-eighth of the total number of seen classes. 
For equal comparison, the baseline of FL-CLSWGAN and ours (VLM) are following exactly the same data split partition in both even and uneven setting.
From Table.~\ref{table:uneven}, we can observe that:(1) The baseline can achieve comparable performance in both even and uneven split. This demonstrates that the extraction of mid-level knowledge is robust to the change in data distribution;
(2) 
\sgg{Our model performance} is superior to FL-CLSWGAN in both even and
uneven \sgg{splits, further validating} the effectiveness and robustness of our method.
}

\begin{figure}[t]
  \centering
  \includegraphics[width=0.38\textwidth]{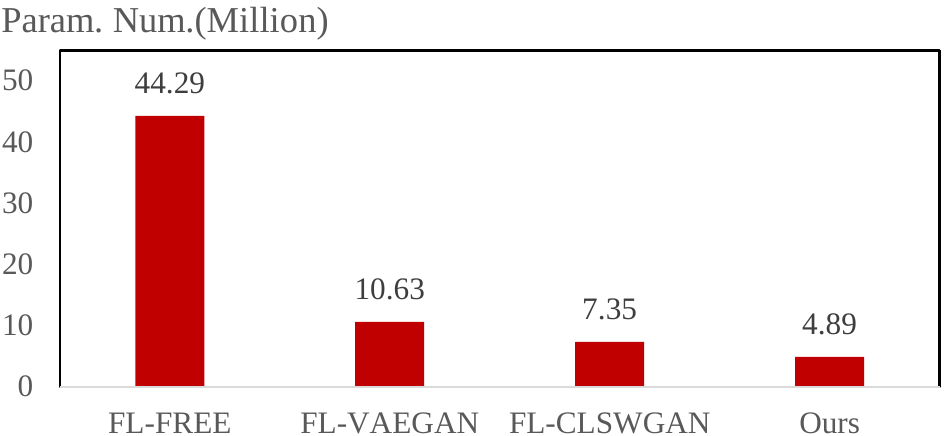}
  \caption{Comparing communication parameters between our mathod and different backbones.
  }
  \label{fig:commParams}
  \end{figure}
  
  \sst{
\textbf{Parameters for communication.}
From Fig.~\ref{fig:commParams}, we can see that: (1) Our approach has the least number of parameters used for the iterative federated communication process comparing with FL-FREE, FL-VAEGAN and FL-CLSWGAN (CLSWGAM+FedAvg in Table.~\ref{table:FZSL}).
(2) This can be attributed to the backbone of our approach CLSWGAN, which is the most lightweight backbone, as well as the SMA that further reduces the communication parameters
(3) The frozen text encoder doesn't join the iterative communication process, we can also implement the text encoder by inference API as in promptFL~\cite{guo2022promptfl}, which can further lighten our model.
As a result, our approach is suitable for federated learning for its efficient communication.
  }

\textbf{Backbone Variations.}
Fig.~\ref{fig:backbone_variation} compares FZSL with different backbone variations. 
We implemented backbones CLSWGAN~\cite{xian2018feature}, FREE~\cite{chen2021free}, VAEGAN~\cite{xian2019f} with FedAvg~\cite{mcmahan2017communication} donated as
FL-CLSWGAN, FL-FREE, FL-VAEGAN respectively, and further utilised our method with SKA(VLM) for comparision. 
We can see that our methods have a clear improvement on all of the three backbones. This validates the effectiveness and generality of our method in FZSL. 

\subsection{Further Analysis and Discussion}
In this section, we conduct the experiments on the improved basedline, namely CLSWGAN+FedAvg+SMA, to evaluate the effect of client number $K$, client fraction $S$ and client local step $E$.

\textbf{Client Number $K$.}
Fig.~\ref{fig:client_num} compares central server aggregation with different numbers of local clients, where 
$K$= 1,2,4 and 10 represent seen classes of the dataset is randomly split to 1,2,4 and 10
clients on average respectively.  
We can see that the FZSL performance decreases when increasing the number of clients, which
implies greater difficulty with larger number of clients with less data variety are used.
But the performance degradation tendencies are stable on CUB and SUN, while that on AWA2 is more significant.

\textbf{Client Fraction $S$.}
Fig.~\ref{fig:client_frac} compares FZSL with different client fractions. 
We can see that a smaller number of fraction is inferior to collaboration with a larger 
fraction of clients, which demonstrates that collaboration among multi-clients can 
further contribute to the generalisation ability of the server model. 

\textbf{Client Local Step $E$.}
Fig.~\ref{fig:local_step} compares FZSL with different client local steps $E$ which influences the communication 
efficiency.
Overall, the performance on different datasets shows relatively stable trends whilst on SUN, the performance decreases when $E$ increases due to the accumulation of biases in local client.

\section{Conclusion}
In this work, we \sgg{proposed} to exploit mid-level semantic
knowledge transfer \sgg{by attributes} for federated learning,
\sgg{from which we introduced} a new Federated Zero-Shot Learning paradigm.
\sst{We \sgg{formulated} a baseline model based on conventional
  zero-shot learning and federated learning. \sgg{We further
    introduced selective module aggregation and foundation model based
    semantic knowledge augmentation} to improve \sgg{the}
  generalisation \sgg{capacity of the} baseline model.
Extensive experiments on five zero-shot learning datasets
\sgg{demonstrated our model's} generalisation \sgg{ability} as well as
\sgg{its effectiveness on privacy preserving}}.


% \begin{acks}
% \end{acks}

\bibliographystyle{ACM-Reference-Format}
\bibliography{sample-base}

% \appendix


\end{document}